%% file: main_full.tex
\documentclass[letterpaper,10pt,conference]{ieeeconf}
\IEEEoverridecommandlockouts
\usepackage{graphicx}
\usepackage[T1]{fontenc}
\usepackage{mathptmx}
\usepackage{amsmath,amssymb}
\usepackage{mathrsfs}
\DeclareMathAlphabet{\mathcal}{OMS}{cmsy}{m}{n}
\usepackage{algorithm,algorithmic}
\usepackage{pgfplots}
\usetikzlibrary{arrows.meta}
\usepackage{placeins}
\usepgfplotslibrary{groupplots}
\pgfplotsset{compat=1.18}
\usepackage{booktabs,multirow}
\newcommand{\method}{WOLF}
\title{\LARGE\bf WOLF: World Model Guided LiDAR Exploration\\with Predictive Frontiers}
\author{Yuyang Tian$^{1,2,*}$, Penghui Yang$^{2,3,*}$, Pengyuan Wu$^{2,4}$, Haoran Yang$^{1,2}$,\\
Chenhui Li$^{2}$, Pengfei Han$^{2}$, Dong Wang$^{2}$, Zhigang Wang$^{2}$,\\
Bin Zhao$^{2,5}$, and Xuelong Li$^{6}$,~\textit{Fellow, IEEE}%
\thanks{$^{*}$Yuyang Tian and Penghui Yang contributed equally to this work.}%
\thanks{$^{1}$University of Science and Technology of China, Hefei, China.}%
\thanks{$^{2}$Shanghai Artificial Intelligence Laboratory, Shanghai, China.}%
\thanks{$^{3}$East China University of Science and Technology, Shanghai, China.}%
\thanks{$^{4}$Zhejiang University, Hangzhou, China.}%
\thanks{$^{5}$Northwestern Polytechnical University, Xi'an, China.}%
\thanks{$^{6}$TeleAI, China.}%
}
\begin{document}
\maketitle
\thispagestyle{empty}
\pagestyle{empty}
\begin{abstract}
LiDAR-based unmanned aerial vehicle (UAV) exploration builds maps by continually selecting where to
observe next. However, decisions based on the measured map provide limited
foresight into spatial continuations behind occlusions, leaving potentially
informative directions unrecognized. We present \method{}, a
world-model-guided framework that predicts future observations to enhance
autonomous exploration. In the training stage, a recurrent world model learns
observation dynamics from exploration trajectories, with recurrent memory
retaining the spatial context needed to interpret partial observations across
successive views. Building on this context, the model combines observation
history with candidate motions during exploration to predict local occupancy
and visibility. To guide further sensing, a predictive frontier generation
mechanism then aligns and fuses these predictions using confidence, branch
agreement, and observation quality to identify promising regions.
The resulting predictive frontiers join measured ones to guide geometric
viewpoint selection and trajectory generation, while new scans update
subsequent predictions. In simulations, our method reduces mean terminal
time by 10.9\% relative to EPIC in Garage at comparable coverage and
increases mean coverage from 42.12\% to 98.35\% in Tunnel. Real-world experiments further
demonstrate onboard deployment of the learned model for online inference
during physical flight.
\end{abstract}
\input{sections/introduction}
\input{sections/related_work}
\input{sections/method}

\input{sections/experiments}

\input{sections/conclusion}
\bibliographystyle{IEEEtran}
\bibliography{references}
\end{document}

%% file: sections/introduction.tex
\section{Introduction}
\label{sec:introduction}

Advances in perception, planning, and control have enabled unmanned aerial vehicles (UAVs) to navigate
complex environments and support inspection, search and rescue, and
three-dimensional mapping~\cite{dang2020graph,schmid2020informative}.
Many of these missions take place in spaces
for which no prior map is available. A robot operating in such settings
must decide where to go, what to observe, and how to extend its knowledge
of the environment as the mission unfolds. Autonomous exploration brings
these decisions together: the UAV must continually discover useful
destinations while building a map through its own actions. Exploration is therefore essential for autonomous missions in
unfamiliar environments.

Geometric exploration methods~\cite{zhou2021fuel,zhang2025falcon,geng2025epic}
organize this process through online
mapping, frontier detection, information gain estimation, and motion
planning. Advances in incremental map management and global planning have
substantially improved exploration efficiency and the scale of deployable
systems. These methods
share a common foundation: they extract candidate observation targets
from measured geometry and organize their visitation according to expected
information gain and travel cost. This foundation connects exploration
decisions to the spatial structure available from onboard sensing.

The remaining challenge is to reason about the value of directions whose
structure is still hidden. A small opening may lead to a large unexplored
region, whereas a direction with substantial immediate gain may soon
terminate. The current map provides only partial evidence for distinguishing
these possibilities. Over a long mission, these choices determine both the information obtained
immediately and the regions that become accessible for subsequent
exploration. Immediate information gain alone therefore gives an incomplete
account of a direction's value. Efficient exploration requires anticipating
how motion will reveal further space, so that decisions reflect the
potential beyond current observations.

World models provide a predictive description of an agent's environment,
learning how observations evolve in response to actions~\cite{hafner2025dreamerv3}.
By integrating observation history, they retain context that is absent
from an individual view and predict possible outcomes before an action is
executed. Recent research has substantially advanced world modelbased flight and
navigation, bringing predictive environmental representations into aerial
decision making~\cite{romero2026dreamtofly,liu2026aird,zhang2026mad}.
For exploration, the central challenge is to make predicted spatial
structure useful for deciding where to observe next. Anticipating the
continuation of a passage or the space behind an opening can reveal
opportunities that are not yet evident in the measured map. Connecting
these predictions to geometric planning enables future observations to
inform the exploration route as the environment is revealed.

We propose \method{}, a world model guided framework for LiDAR-based aerial
exploration. Its central idea is to bring learned spatial prediction into
the formation of exploration targets. A recurrent world model learns from
trajectories across multiple environments and predicts local observations
along candidate motions. We convert these predictions into
\emph{predictive frontiers}, which expose potential spatial continuations
as explicit targets for further sensing. Planning over both measured and
predictive frontiers allows the UAV to consider opportunities suggested
by experience alongside those already evident in the map.

The framework integrates spatial prediction into geometric planning.
The world model informs which additional regions merit observation, while
measured geometry supports viewpoint selection, reachability analysis,
and trajectory generation. Executed motion produces new observations
that update the model state and revise the available targets. Through
this interaction, learned experience influences the exploration route,
and incoming measurements continually refine its direction. Predictive
frontiers provide the connection from local spatial anticipation to the
progress of an extended exploration mission.

Experiments across complementary environments demonstrate that this
integration improves exploration capability: the UAV reaches broad
coverage more efficiently and sustains exploration into a larger portion
of an unfamiliar environment. The results support using spatial prediction
to guide more effective exploration routes and discover useful directions
as the mission progresses. Real-world flight further shows that predicted
spatial continuations are supported by subsequent LiDAR observations,
connecting the learned predictions to space encountered during physical
exploration.

The main contributions of this work are:
\begin{itemize}
    \item We introduce an exploration oriented world model that learns
    spatial observation dynamics from experience across environments,
    enabling predictions of sensing outcomes along candidate directions.
    \item We develop a predictive frontier generation mechanism that combines
    spatial prediction with geometric exploration, allowing predicted
    structure to contribute to effective route planning.
    \item Experimental results show improved exploration efficiency and
    coverage in simulation and successful onboard model inference in
    real-world flights.
\end{itemize}

%% file: sections/related_work.tex
\section{Related Work}
\label{sec:related_work}

\subsection{Geometric Aerial Exploration}

Frontier and next-best-view methods select sensing locations that reveal
unknown space. NBVP~\cite{bircher2016nbvp} evaluates informative branches in a receding-horizon
sampling tree. Subsequent work develops efficient
frontier extraction, history-aware recovery, and informative trajectory
optimization~\cite{cieslewski2017rapid,witting2018history,dai2020fastfrontier,
schmid2020informative}. Local information gain can be combined with global
connectivity to support exploration over larger environments, as in AEP~\cite{selin2019aep} and
graph-based exploration~\cite{dang2020graph}.
FUEL~\cite{zhou2021fuel} maintains an incremental frontier structure and coordinates global
viewpoint ordering with local trajectory planning.
Bubble Explorer~\cite{tang2023bubble} constructs occlusion-free spheres for viewpoint
generation, while next-best-trajectory planning jointly
considers motion and observation utility~\cite{lindqvist2024nbt}.
FALCON~\cite{zhang2025falcon} introduces online coverage paths to guide exploration across
regions. EPIC~\cite{geng2025epic} uses point-surface observation quality
and an incrementally maintained point cloud topological graph for
large-scale LiDAR exploration. These developments address
complementary aspects of exploration, including target extraction, map
representation, global organization, and trajectory execution.

\subsection{Prediction-Assisted Exploration}

Learned spatial predictions provide additional information for evaluating
where a robot should observe. Learned Map Prediction~\cite{shrestha2019map} completes unseen
regions of a two-dimensional map and uses those completions to guide mobile
robot exploration. SEER~\cite{tao2023seer} predicts local unseen occupancy
and estimates viewpoint information gain for aerial
exploration. PredRecon~\cite{feng2023predrecon} uses coarse surface predictions to
organize viewpoints for autonomous reconstruction.
At a larger spatial scale, MapEx~\cite{ho2025mapex} samples global map completions and evaluates
observation utility through probabilistic information
gain. PIPE~\cite{baek2025pipe} integrates predicted information gain along
candidate paths to account for overlapping observations and gain
estimation. Together, these methods investigate how
spatial prediction, visibility, and uncertainty inform sensing decisions.

\subsection{World Models for Aerial Autonomy}

World models learn predictive states from observation and action sequences.
DreamerV3~\cite{hafner2025dreamerv3} combines recurrent latent dynamics with policy learning in imagined
trajectories across diverse tasks.
Model-based learning for quadrotors has also addressed low-level control and
latent vehicle dynamics~\cite{lambert2019lowlevel,becker2020learning}.
More recent aerial systems use world models to connect visual observations,
temporal context, and flight decisions. Dream to Fly~\cite{romero2026dreamtofly} learns racing policies
from RGB observations and evaluates transfer through a hardware-in-the-loop
quadrotor setup. SkyDreamer~\cite{verraest2025skydreamer} investigates
interpretable representations for vision-based drone
racing. AirDreamer~\cite{liu2026aird} learns an environment
representation from depth and proprioception and trains a navigation policy
with sparse rewards. MAD~\cite{zhang2026mad} supervises recurrent states with
local occupancy, visibility, and proprioception and evaluates their use
across policy learning algorithms and flight tasks.
FlowPilot~\cite{wang2026flowpilot} jointly models future depth observations and executable
trajectories for onboard navigation.

%% file: sections/method.tex
% Replacement Method section for the existing WOLF manuscript.
% Keep the original bibliography entries hafner2025dreamerv3 and geng2025epic.
% Required packages: amsmath, amssymb, graphicx, algorithm, algorithmic.
% The figure path is retained from the supplied manuscript.

\section{Method}
\label{sec:method}

\subsection{Problem Definition}

We consider LiDAR-based exploration of an initially unknown environment. At step $t$, a UAV acquires a scan $o_t$ at pose $x_t=(p_t,\psi_t)$, where $p_t\in\mathbb{R}^3$ is the position and $\psi_t$ is the yaw angle, and updates the accumulated measured map $G_t$. The objective is to select sensing poses that increase surface coverage while accounting for travel cost and motion feasibility. Occlusions limit the geometry available in $G_t$, leaving the spatial extent that a candidate motion may reveal only partially known.

\begin{figure*}[t]
    \centering
    \includegraphics[width=0.97\linewidth]{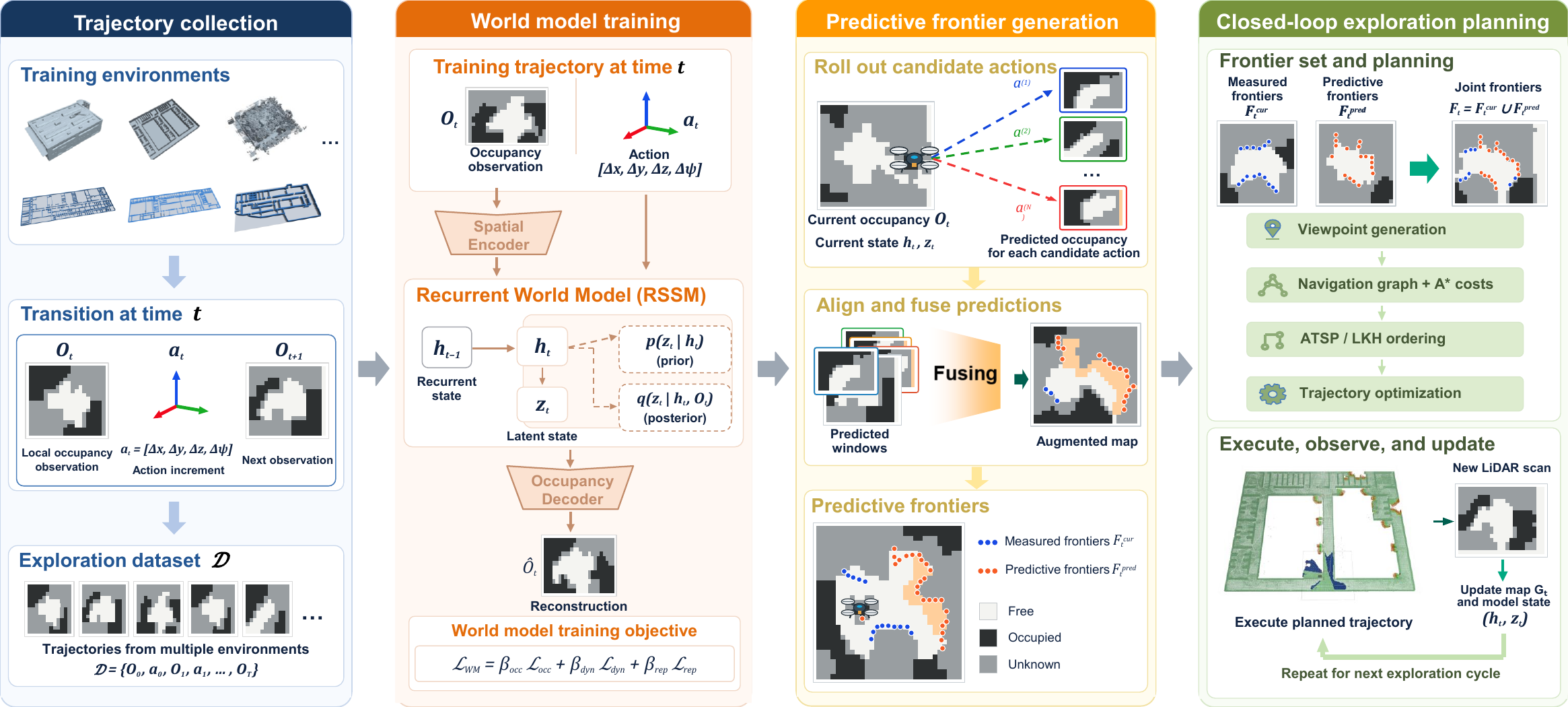}
    \caption{\textbf{Overview of WOLF.} From left to right, exploration trajectories collected across environments provide occupancy--action sequences for offline training of a recurrent world model. During exploration, the model combines observation history with candidate motions to predict future local occupancy observations. These predictions are spatially aligned and fused using confidence, branch agreement, and observation quality to extract predictive frontiers. Measured and predictive frontiers jointly guide viewpoint generation and trajectory planning, while the measured map supplies reachability and collision checks. Executed motion and new LiDAR scans update the map and model state for the next exploration cycle.}
    \label{fig:wm_epic_overview}
\end{figure*}

To predict what lies beyond the observed geometry, WOLF uses action conditioned observation prediction to identify additional sensing targets (Fig.~\ref{fig:wm_epic_overview}). A recurrent world model is trained offline on exploration trajectories to capture how local observations change with UAV motion. During exploration, the model integrates the latest scan into its recurrent state and predicts future local observations under candidate pose increments. These predictions are aligned and fused to extract \emph{predictive frontiers}, which are planned jointly with frontiers derived from measurements. The world model supplies additional regions to observe, while the geometric planner~\cite{geng2025epic} uses measured geometry to select feasible viewpoints and generate trajectories. Executed motion and new scans update the model state and map for the next planning cycle.

\subsection{World Model Training}
\label{sec:spatial_world_model}

The world model represents local observations and motion between scans in a common world-aligned frame. A recurrent latent state retains spatial context across successive views and supports both observation updates and action conditioned prediction.

\paragraph{Observation space}
We rasterize each LiDAR scan into a local bird's-eye-view occupancy map:
\begin{equation}
    O_t=R(o_t,x_t)\in C^{H\times W},\qquad
    C=\{\mathrm{unk},\mathrm{free},\mathrm{occ}\},
    \label{eq:observation_action}
\end{equation}
where $R$ considers LiDAR returns within a vertical slab around the flight region. Ray traversed cells are labeled free, return endpoints occupied, and cells without measurement evidence unknown. Occupied labels take precedence when projected rays overlap. The window is centered at the UAV position, with its horizontal axes fixed to the world frame. The three classes encode both local geometry and scan visibility, including the free unknown boundaries used for exploration. Each input contains only the current scan; preceding observations are represented by the recurrent state.

\paragraph{Action space}
An action denotes the realized pose increment between consecutive scans, rather than a low-level control command:
\begin{equation}
    a_t=\begin{bmatrix}\Delta p_t\\\Delta\psi_t\end{bmatrix},\qquad
    \Delta p_t=p_{t+1}-p_t,
    \label{eq:action_space}
\end{equation}
where $\Delta p_t=[\Delta x_t^w,\Delta y_t^w,\Delta z_t^w]^\top$ is expressed in the world frame. The yaw increment is wrapped to $[-\pi,\pi]$:
\begin{equation}
    \Delta\psi_t=\operatorname{atan2}\!\left(
    \sin(\psi_{t+1}-\psi_t),\cos(\psi_{t+1}-\psi_t)\right).
    \label{eq:yaw_increment}
\end{equation}
Using realized increments pairs each observation transition with the motion that actually occurred, including partial progress toward a selected viewpoint. The same representation is used for candidate motions at inference.

\paragraph{Recurrent world model}
We adopt the recurrent state space model (RSSM) of DreamerV3~\cite{hafner2025dreamerv3}, with a deterministic recurrent state $h_t$ and a stochastic categorical state $z_t$. The model comprises a spatial encoder, recurrent latent dynamics, and an occupancy decoder, jointly parameterized by $\theta$. Reward and continuation prediction heads are omitted, as exploration decisions are made by the geometric planner. The encoder and decoder use convolutional neural networks (CNNs), while the prior and posterior use multilayer perceptrons (MLPs).

The spatial encoder extracts spatial features from the current observation. 
\begin{equation}
    \text{Encoder:}\qquad e_t=E_\theta(O_t),
    \label{eq:spatial_encoder}
\end{equation}
where $e_t$ provides the local geometric evidence used to infer the latent state. 
% Spatial downsampling aggregates the arrangement of occupied surfaces and free-space boundaries within the observation window.

The recurrent transition propagates the preceding state under the executed action, and the prior and posterior infer the categorical state without and with the current observation, respectively:
\begin{align}
    \text{Sequence model:}\quad &h_t=f_\theta(h_{t-1},z_{t-1},a_{t-1}),
        \label{eq:rssm}\\
    \text{Prior:}\quad &\hat z_t\sim p_\theta(z_t\mid h_t),
        \label{eq:latent_prior}\\
    \text{Posterior:}\quad &z_t\sim q_\theta(z_t\mid h_t,e_t).
        \label{eq:observation_posterior}
\end{align}
During training and online state estimation, the posterior incorporates newly observed geometry and supplies the state for the next transition. For a hypothetical motion, the prior supplies the predicted state before the corresponding scan is available. Both operations use the same recurrent dynamics.

The decoder maps the joint state to a categorical distribution over occupancy classes at each cell $u$:
\begin{equation}
    \text{Decoder:}\qquad  P_t(u,c)=D_\theta(h_t,z_t)_{u,c},\qquad c\in C.
    \label{eq:occupancy_decoder}
\end{equation}
It projects the concatenated latent features into a spatial feature map, followed by transposed convolutions and a softmax over classes. Decoding a posterior state reconstructs the acquired observation; decoding a motion conditioned prior state predicts a future observation. The decoded free and unknown probabilities expose the anticipated observation boundaries used to generate predictive frontiers.

\paragraph{Training objective}
We train the model on complete observation and action trajectories. At each valid time step, the posterior reconstruction is supervised by a class weighted occupancy loss:
\begin{equation}
    \mathcal{L}_{\mathrm{occ}}
    =-\mathbb{E}_{t}\left[\sum_{u\in U}\sum_{c\in C}
    w_c\mathbf{1}[O_t(u)=c]\log P_t(u,c)\right],
    \label{eq:occupancy_loss}
\end{equation}
where $U$ is the set of grid cells and $w_c$ compensates for class imbalance. Following DreamerV3~\cite{hafner2025dreamerv3}, two gradient-separated KL terms align the prior and posterior:
% \begin{align}
%     \ell_{\mathrm{dyn},t}
%     &=\max\!\big(\lambda,\nonumber\\
%     &\quad D_{\mathrm{KL}}[\operatorname{sg}(q_\theta(z_t\mid h_t,e_t))
%     \|p_\theta(z_t\mid h_t)]\big),
%     \label{eq:latent_loss}\\
%     \ell_{\mathrm{rep},t}
%     &=\max\!\big(\lambda,\nonumber\\
%     &\quad D_{\mathrm{KL}}[q_\theta(z_t\mid h_t,e_t)
%     \|\operatorname{sg}(p_\theta(z_t\mid h_t))]\big).
%     \label{eq:representation_loss}
% \end{align}
\begin{align}
    \mathcal{L}_{\mathrm{dyn}}&=\mathbb{E}_t\big[\max\big(\lambda,\nonumber\\[-2pt]
    &\quad D_{\mathrm{KL}}[\operatorname{sg}(q_\theta(z_t\mid h_t,e_t))
    \|p_\theta(z_t\mid h_t)]\big)\big],
    \label{eq:latent_loss}\\
    \mathcal{L}_{\mathrm{rep}}&=\mathbb{E}_t\big[\max\big(\lambda,\nonumber\\[-2pt]
    &\quad D_{\mathrm{KL}}[q_\theta(z_t\mid h_t,e_t)
    \|\operatorname{sg}(p_\theta(z_t\mid h_t))]\big)\big].
    \label{eq:representation_loss}
\end{align}
The dynamics loss trains the prior to match the observation-conditioned posterior, while the representation loss regularizes the posterior toward the predictive prior. Here, $\operatorname{sg}(\cdot)$ denotes stop-gradient and $\lambda=1$ is the free-nats threshold. The overall objective is
\begin{equation}
    \begin{split}
    \mathcal{L}_{\mathrm{WM}}=\mathbb{E}_{B}\big[&
    \beta_{\mathrm{occ}}\mathcal{L}_{\mathrm{occ}}
    +\beta_{\mathrm{dyn}}\mathcal{L}_{\mathrm{dyn}}\\
    &+\beta_{\mathrm{rep}}\mathcal{L}_{\mathrm{rep}}\big],
    \end{split}
    \label{eq:world_model_loss}
\end{equation}
where the expectation averages over trajectory batches $B$ and their valid time steps, with latent states sampled from the posterior during sequence unrolling. We use $\beta_{\mathrm{occ}}=1$, $\beta_{\mathrm{dyn}}=1$, and $\beta_{\mathrm{rep}}=0.1$.

\subsection{Predictive Frontier Generation}

We convert candidate conditioned predictions into additional free unknown boundaries. The procedure comprises batched prediction, confidence gated fusion, and budgeted frontier extraction.

\paragraph{Candidate conditioned prediction}
At each planning step, the latest scan updates the posterior state $(h_t,z_t)$. The planner constructs candidate sensing actions $A_t=\{a_t^i\}_{i=1}^{N}$ from frontier viewpoints and directional samples. Before prediction, each candidate translation $\Delta p_i$ is rescaled by
\begin{equation}
    \alpha_i=\min\!\left(1,
    \frac{d_{\max}}{\max(\|\Delta p_i\|,\epsilon)}\right),
    \label{eq:candidate_translation_scale}
\end{equation}
and its yaw increment is clipped to $[-\pi,\pi]$, where $d_{\max}=3\,\mathrm{m}$ and $\epsilon=10^{-6}$ avoids division by zero. The rescaled translation $\alpha_i\Delta p_i$ preserves the direction toward a distant viewpoint while bounding the motion by $d_{\max}$. These bounds keep candidate motions at the short transition scale represented in the training trajectories. Below, $a_t^i$ denotes the bounded pose increment.

All candidates branch from the same current state and undergo one prior transition, computed in a batch:
\begin{equation}
    \begin{aligned}
    \hat h_{t+1}^i&=f_\theta(h_t,z_t,a_t^i),\\
    \hat z_{t+1}^i&\sim p_\theta(z_{t+1}\mid\hat h_{t+1}^i),\\
    P_{t+1}^i(u,c)&=D_\theta(\hat h_{t+1}^i,\hat z_{t+1}^i)_{u,c}.
    \end{aligned}
    \label{eq:branches}
\end{equation}
Each predicted window is centered at its candidate endpoint. Because the raster axes are world aligned, the windows are translated onto a common local grid for fusion at corresponding physical locations. In the following, $P_{t+1}^i$ denotes the aligned prediction, and a branch contributes only within its valid window.
\begin{algorithm}[]
\caption{Predictive Frontier Generation}
\label{alg:predictive_frontier}
\begin{algorithmic}[1]
\REQUIRE Local measured raster $L_t$, measured frontiers $F_t^{\mathrm{cur}}$, aligned predictions $\{P_{t+1}^i\}$, history mask $H_t$, and poorly observed neighborhood $R_t$
\ENSURE Predictive frontiers $F_t^{\mathrm{pred}}$
\STATE $\widetilde L_t\gets L_t$, $Q\gets\emptyset$
\FOR{each unknown cell $u$ in $L_t$}
    \STATE Compute supported scores and counts using~\eqref{eq:confidence_support}
    \IF{at least one known class is supported}
        \STATE Select $c^*=\arg\max_c s_c(u)$ over supported classes
        \IF{no cross-class conflict and~\eqref{eq:acceptance} holds}
            \STATE Add $(u,c^*,s_{c^*}(u),\mathrm{priority}(u))$ to $Q$
        \ENDIF
    \ENDIF
\ENDFOR
\STATE Compute the spatial budget $B_t$ using~\eqref{eq:spatial_budget}
\STATE Sort $Q$ by observation priority, then confidence
\STATE Write the first $\min(B_t,|Q|)$ entries to $\widetilde L_t$; record their cell set $S_t$
\STATE Extract connected components of~\eqref{eq:frontier_mask}
\STATE Remove components duplicating $F_t^{\mathrm{cur}}$
\STATE Attach contributing branches predicting free space to each retained component
\RETURN $F_t^{\mathrm{pred}}$
\end{algorithmic}
\end{algorithm}

\paragraph{Confidence gated fusion}
Fusion starts from a local measured raster $L_t$ initialized from the current observation. Let $H_t$ denote cells unknown in the accumulated observation history, and let $R_t$ be a two cell neighborhood around surfaces marked as poorly observed by the geometric mapping module. These sets identify new regions and unresolved boundaries, respectively. Predictions are added only to unknown cells in $L_t$; its measured labels remain unchanged.

At cell $u$, each valid branch proposes its most probable class, $c_i(u)=\arg\max_{c\in C}P_{t+1}^i(u,c)$. For each known class $c\in\{\mathrm{free},\mathrm{occ}\}$, let $I_c(u)$ contain the branches proposing $c$. Its confidence score and support count are
\begin{equation}
    s_c(u)=\max_{i\in I_c(u)}P_{t+1}^i(u,c),\qquad
    n_c(u)=|I_c(u)|.
    \label{eq:confidence_support}
\end{equation}
Classes without supporting branches are excluded. Among the supported classes, we select $c^*=\arg\max_c s_c(u)$. If both classes receive support and $|s_{\mathrm{free}}(u)-s_{\mathrm{occ}}(u)|\leq\delta$, the cell remains unknown. Otherwise, the selected class is eligible for fusion when
\begin{equation}
    \begin{aligned}
    \mathrm{accept}(u)={}&[s_{c^*}(u)\geq\tau_{c^*}]\;\lor\\
    &\big[u\in R_t\;\land\;n_{c^*}(u)\geq k\\
    &\quad\land\;s_{c^*}(u)\geq\tau^{\mathrm{poor}}_{c^*}\big].
    \end{aligned}
    \label{eq:acceptance}
\end{equation}
Thus, a strict confidence threshold admits a prediction directly, while a relaxed threshold near poorly observed surfaces additionally requires support from at least $k$ candidate branches. We set $(\tau_{\mathrm{free}},\tau_{\mathrm{occ}})=(0.75,0.97)$, $(\tau_{\mathrm{free}}^{\mathrm{poor}},\tau_{\mathrm{occ}}^{\mathrm{poor}})=(0.68,0.94)$, $k=2$, and $\delta=0.05$.

\paragraph{Budgeted frontier extraction}
Eligible cells are ranked first by observation priority and then by confidence, giving precedence to cells in $H_t\cup R_t$. With $K_t$ measured known cells in $L_t$, the number of added cells is limited to
\begin{equation}
    B_t=\max\{0,\min(HW,\lfloor\rho K_t\rfloor)-K_t\}.
    \label{eq:spatial_budget}
\end{equation}
Here, $1\leq\rho\leq2$. The selected cell set $S_t$, with $|S_t|\leq B_t$, forms a temporary augmented raster $\widetilde L_t$. This budget bounds the augmented known area relative to the current measured footprint.

Predictive frontiers are extracted from added free cells that have observation priority and border unknown space:
\begin{equation}
    \begin{aligned}
    &M_t=\{u\in S_t\cap(H_t\cup R_t):\\
    &\qquad\widetilde L_t(u)=\mathrm{free},\\
    &\qquad\exists v\in N(u),\;
    \widetilde L_t(v)=\mathrm{unk}\},
    \end{aligned}
    \label{eq:frontier_mask}
\end{equation}
where $N(u)$ denotes the neighboring cells of $u$. Connected components of $M_t$ form the predictive frontier set $F_t^{\mathrm{pred}}$ after components duplicating measured frontiers $F_t^{\mathrm{cur}}$ are removed. Each region retains the candidate branches contributing its evidence of free space, preserving the associated sensing directions at the planning interface. Algorithm~\ref{alg:predictive_frontier} summarizes the procedure.

\subsection{Closed-Loop Exploration Planning}

The planner operates on the joint frontier set $F_t=F_t^{\mathrm{cur}}\cup F_t^{\mathrm{pred}}$. A predictive frontier specifies a region to observe; viewpoint generation selects a pose from which that region can be sensed. Both frontier types pass through the same geometric viewpoint generation pipeline. The temporary raster $\widetilde L_t$ supplies predictive frontier geometry, while the accumulated measured map $G_t$ supplies visibility, obstacle clearance, reachability, and collision checks. Candidates overlapping existing viewpoints or recently visited targets are removed before joint planning.

Following EPIC~\cite{geng2025epic}, retained viewpoints are connected to the navigation graph. A* search provides collision-free graph paths and travel costs between viewpoints. Together with connections from the current UAV pose, these costs form a directed cost matrix. We formulate joint viewpoint ordering as an asymmetric traveling salesman problem (ATSP) and use the Lin Kernighan Helsgaun (LKH) solver to obtain a low-cost visitation order. The corresponding graph paths define a global guidance route, which local optimization converts into an executable trajectory under geometric and dynamic constraints.

After execution, the realized pose increment and next scan update the persistent posterior, while the scan also updates $G_t$. Temporary prediction branches are discarded, and candidate motions are regenerated from the updated state. The resulting loop uses predicted spatial continuations to propose additional views and newly measured geometry to revise subsequent decisions. When $F_t^{\mathrm{pred}}$ is empty, planning proceeds with measured frontiers.

%% file: sections/experiments.tex
\section{Simulations and Experiments}
\label{sec:experiments}
Our evaluation examines whether learned spatial foresight improves the
outcome of an exploration mission. We compare the time required to achieve
high coverage and the extent of exploration in an unseen environment, then
examine spatial predictions and their use during physical flight.

\subsection{Experimental Setup}

\input{tables/exploration_results_full_stats}

\begin{figure*}[t]
    \centering
    \includegraphics[width=0.97\linewidth]{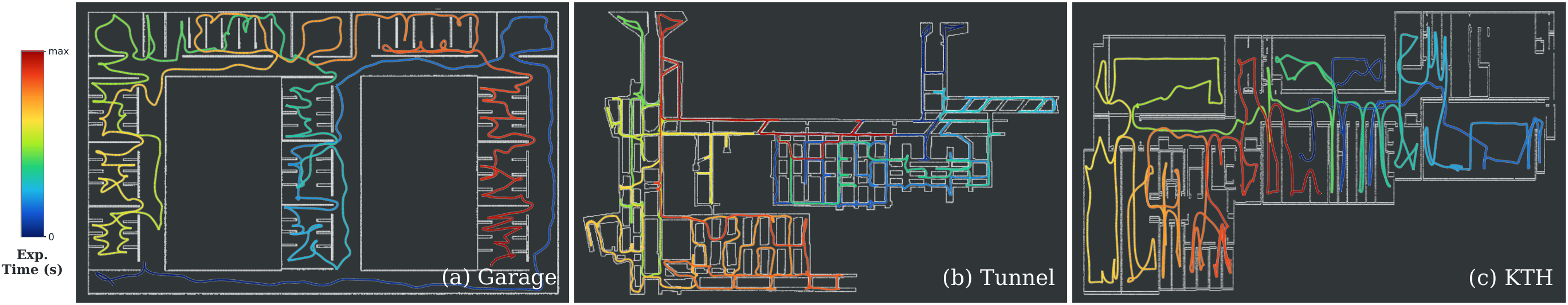}
    \caption{Representative \method{} trajectories in Garage, Tunnel, and
    KTH, colored by elapsed mission time. The trajectories illustrate the
    spatial extent of exploration across open areas, extended passages,
    and connected rooms.}
    \label{fig:three_scene_trajectories}
\end{figure*}

\begin{figure*}[t]
    \centering
    \includegraphics[width=0.97\linewidth]{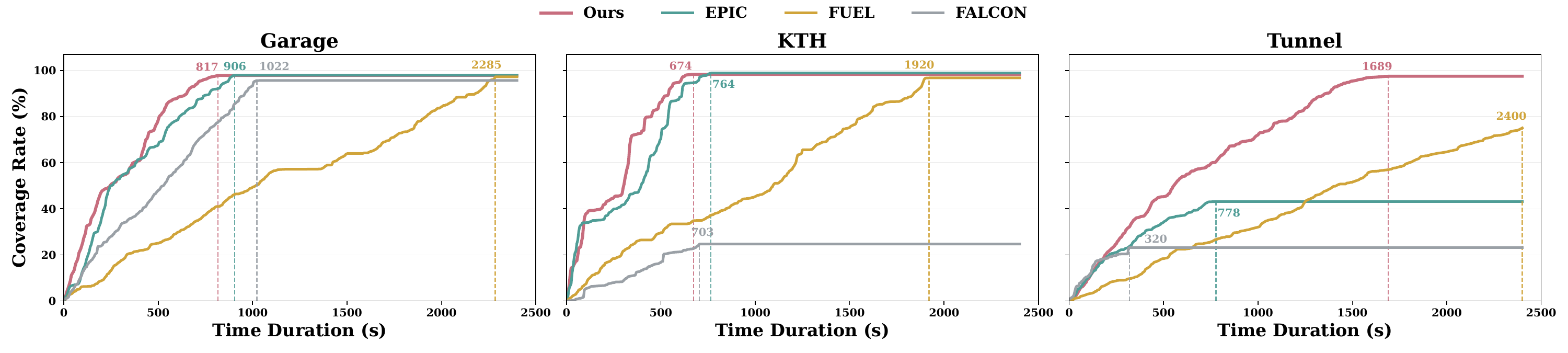}
    \caption{The variation of coverage rate with flight time during
    exploration in Garage, KTH, and Tunnel (left to right).}
    %\vspace{-10pt}
    \label{fig:three_scene_coverage}
\end{figure*}

We use MARSIM~\cite{kong2023marsim} as the simulation platform for
collecting exploration experience. The resulting dataset comprises 10K
trajectories and 1M observation--action transitions collected across
33 environment maps. Our model is trained offline on complete
trajectories for 200 epochs using two NVIDIA RTX 4090 GPUs for 60 hours. Each LiDAR
observation is represented as a $96\times96$ occupancy grid covering a
$50\,\mathrm{m}\times50\,\mathrm{m}$ local region. We evaluate exploration performance against FUEL~\cite{zhou2021fuel},
FALCON~\cite{zhang2025falcon}, and EPIC~\cite{geng2025epic} in Garage,
KTH, and Tunnel, with a common maximum velocity of $4\,\mathrm{m/s}$.
Coverage rate follows EPIC~\cite{geng2025epic}.
Each baseline runs 10 times per environment with a $2400\,\mathrm{s}$ time limit.
Garage and KTH are represented in the training corpus, whereas Tunnel is
held out from training to assess generalization to an unseen layout.

\subsection{Exploration Efficiency and Completeness}
Table~\ref{tab:three_scene_results_full} shows that \method{} combines
high coverage with efficient exploration across the three environments.
In Garage and KTH, it achieves the shortest mean terminal time and path
length among methods with mean coverage above 95\%. In the unseen Tunnel,
it is the only evaluated method to exceed this coverage threshold on
average, demonstrating a substantial improvement in exploration completeness.

In Garage, \method{} reaches 98.22\% mean coverage while reducing mean
terminal time by 10.9\% and path length by 10.8\% relative to EPIC.
Compared with FUEL, it completes exploration in 58.4\% less time at
slightly higher coverage. The shorter route and mission duration show
that the efficiency gain accompanies broad observation of the environment.
The same benefit extends to KTH: \method{} maintains 98.31\% mean
coverage, close to EPIC's 98.61\%, while reducing mean terminal time by
8.5\% and path length by 5.7\%. Thus, the framework improves exploration
efficiency in both open passages and connected rooms.

The improvement is most pronounced in Tunnel, where \method{} reaches
98.35\% mean coverage, exceeding FUEL, the strongest coverage baseline,
by 38.6 percentage points and covering more than twice the surface reached
by EPIC. Coverage remains above 96.9\% in every reported \method{} run.
The longer route and mission relative to EPIC accompany exploration of a
much larger portion of the environment. These results demonstrate that
predictive frontiers help sustain useful exploration directions in an
unfamiliar layout, extending the mission toward nearly complete coverage.

Fig.~\ref{fig:three_scene_trajectories} illustrates the spatial scope of
these missions: the routes traverse Garage's open passages, Tunnel's
extended branches, and KTH's connected rooms. Together with the coverage rate
curves shown in Fig.~\ref{fig:three_scene_coverage}, these trajectories
connect the quantitative gains to the exploration across different
spatial structures. The combined evidence supports predictive frontiers
as a means of improving both the efficiency and coverage of exploration.
\begin{table}[t]
\centering
\caption{Average time consumption of each module.}
\label{tab:module_timing}
\scriptsize
\setlength{\tabcolsep}{1.5pt}
\begin{tabular*}{\columnwidth}{@{\extracolsep{\fill}}ccccc@{}}
\toprule
Scene & \shortstack{World model\\response (ms)} & \shortstack{Predictive\\viewpoints (ms)} & \shortstack{Path cost\\matrix (ms)} & \shortstack{TSP\\solving (ms)} \\
\midrule
Garage & 28.84 & 0.98 & 16.43 & 4.21 \\
KTH & 19.85 & 0.96 & 3.26 & 2.79 \\
Tunnel & 31.06 & 1.75 & 10.79 & 3.58 \\
\bottomrule
\end{tabular*}
\end{table}

The computational results in Table~\ref{tab:module_timing} complement
these exploration gains. Although world model prediction is the most time-consuming component, its modest runtime allows spatial prediction
to be integrated without compromising real-time exploration.

\subsection{Spatial Prediction Analysis}

\begin{figure}[htbp]
    \centering
    \includegraphics[width=0.94\linewidth]{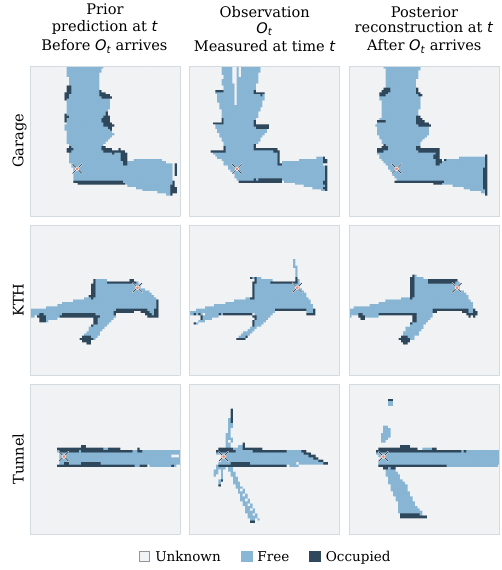}
    \caption{Prior predictions, observations, and posterior reconstructions
    at time $t$ (left to right). The prior predicts $O_t$ before it arrives;
    the posterior reconstructs $O_t$ after incorporating it.}
    \label{fig:occupancy_prior_posterior}
    %\vspace{-10pt}
\end{figure}

Fig.~\ref{fig:occupancy_prior_posterior} compares prior predictions,
measured observations, and posterior reconstructions across the three scenes.
The prior anticipates the dominant free space structure before the next
scan arrives. In Garage, it preserves the extended passage and surrounding
occupied boundaries; in KTH, it captures the arrangement of connected
spaces around the central junction. These predictions provide spatially
structured candidates for further sensing, allowing the planner to consider
where an observed passage or opening may continue.

The unseen Tunnel illustrates both transfer of this structural prediction
and refinement through observation. The prior retains the main passage,
while the posterior recovers the oblique branch more clearly after
incorporating the scan. Across the examples, posterior updates refine
local boundaries and scene-specific details while preserving the dominant
layout. This relationship is useful for exploration: the prior supplies
an initial hypothesis of spatial continuation, and the posterior integrates
the evidence revealed by the resulting motion.

For frontier generation, the relevant information is the arrangement of
predicted free space relative to unknown regions. A coherent continuation
can expose an additional observation target near the boundary of the
current view. The examples show how the learned representation provides
this information across different layouts, complementing the mission-level
improvements in efficiency and explored extent. Prediction thus contributes
to exploration through the sensing opportunities it reveals and their
subsequent refinement by measurements.

\begin{table}[htbp]
    \centering
    \caption{Prediction of the next observation on a fixed trajectory in the
    KTH environment.}
    \label{tab:room4m_prediction_ablation}
    \footnotesize
    \setlength{\tabcolsep}{2pt}
    \begin{tabular*}{\columnwidth}{@{\extracolsep{\fill}}lccc@{}}
        \toprule
        Method & \shortstack{Action-\\conditioned} & \shortstack{mIoU (\%)} &
        \shortstack{Improvement over\\MapEx (pp)} \\
        \midrule
        MapEx & No & 51.36 & -- \\
        Ours w/o action & No & 46.36 & $-5.00$ \\
        \textbf{Ours (full)} & Yes & \textbf{65.54} &
        $\mathbf{+14.18}$ \\
        \bottomrule
    \end{tabular*}
\end{table}

\begin{figure}[]
    \centering
    \includegraphics[width=\linewidth]{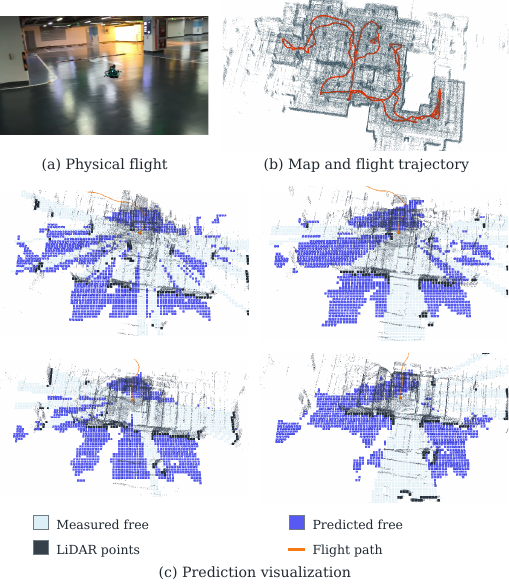}
    \caption{Real-world exploration: (a) platform; (b) map and trajectory;
    (c) successive inference results.}
    \label{fig:real_world_exploration}
    %\vspace{-10pt}
\end{figure}
\FloatBarrier

\subsection{Ablation Study}

We compare action-conditioned observation prediction with 
MapEx~\cite{ho2025mapex} at different steps of a fixed
KTH trajectory. All methods receive the same observation history and are
evaluated against the next measured local observation. MapEx's completed
global map is projected into the same local grid using the next pose
and sensor model; our model predicts that observation directly from its
RSSM state and the executed action.

As shown in Table~\ref{tab:room4m_prediction_ablation}, our full model
achieves an absolute mIoU gain of 14.18\% over MapEx. Without action
input, mIoU drops to 46.36\%. Together, these results demonstrate the advantage
over global map completion for predicting the next observation and highlight
the importance of modeling how robot motion changes the observed space.

\subsection{Real-World Exploration}

\label{sec:real_world}
We deploy the framework on a quadrotor equipped with a $360^{\circ}$
LiDAR and an NVIDIA Jetson Orin NX 16 GB onboard computer. Experiments are conducted in an
underground parking garage, where rows of columns and walls divide open
passages and create occluded regions. The recorded configuration bounds
the exploration area to $80\,\mathrm{m}\times45\,\mathrm{m}$ and sets the
maximum velocity limit to $1.0\,\mathrm{m/s}$.
The real-world exploration finishes after
$480.2\,\mathrm{s}$, while onboard global planning takes
$260.0\,\mathrm{ms}$ on average. Fig.~\ref{fig:real_world_exploration} presents the
physical platform, an onboard prediction visualization, and the RViz map with the flight trajectory. These
results demonstrate operation of the prediction-guided exploration
framework on the onboard platform in a physically occluded environment.

%% file: tables/exploration_results_full_stats.tex
% Drop-in alternative to exploration_results_mean_std.tex.
% Do not input both files simultaneously because they intentionally share a label.
\begin{table*}[t]
    \centering
    \caption{Results of simulations in different environments.}
    \label{tab:three_scene_results_full}
    \scriptsize
    \setlength{\tabcolsep}{1.7pt}
    \resizebox{\textwidth}{!}{%
    \begin{tabular}{ccrrrrrrrrrrrrrrrr}
        \toprule
        \multirow{2}{*}{Scene} & \multirow{2}{*}{Method} & \multicolumn{4}{c}{Coverage (\%) $\uparrow$} &
        \multicolumn{4}{c}{Terminal time (s) $\downarrow$} &
        \multicolumn{4}{c}{Path (m) $\downarrow$} &
        \multicolumn{4}{c}{Avg. Vel. (m/s) $\uparrow$} \\
        \cmidrule(lr){3-6}\cmidrule(lr){7-10}\cmidrule(lr){11-14}\cmidrule(lr){15-18}
        & & Mean & Std. & Min. & Max. & Mean & Std. & Min. & Max. &
        Mean & Std. & Min. & Max. & Mean & Std. & Min. & Max. \\
        \midrule
        \multirow{4}{*}{Garage} & FUEL
          & 97.11 & 0.12 & 97.02 & 97.28
          & 2024.8 & 268.0 & 1656.2 & 2285.4
          & 3017.4 & 353.6 & 2537.1 & 3378.0
          & 1.49 & 0.03 & 1.47 & 1.53 \\
        & FALCON\textsuperscript{\P}
          & 76.36 & 18.30 & 41.55 & 95.66
          & 748.7 & 211.7 & 375.6 & 1022.5
          & 1105.5 & 319.2 & 569.6 & 1560.1
          & 1.48 & 0.07 & 1.34 & 1.53 \\
        & EPIC
          & 97.88 & 0.13 & 97.63 & 98.12
          & 944.4 & 70.7 & 801.8 & 1044.7
          & 2965.3 & 221.8 & 2646.6 & 3242.6
          & \textbf{3.15} & 0.20 & 2.92 & 3.38 \\
        & \textbf{Ours}
          & \textbf{98.22} & 0.11 & 97.97 & 98.34
          & \textbf{841.5} & 58.2 & 736.4 & 934.1
          & \textbf{2644.5} & 153.3 & 2338.8 & 2943.7
          & \textbf{3.15} & 0.16 & 2.82 & 3.37 \\
        \midrule
        \multirow{4}{*}{KTH} & FUEL\textsuperscript{\P}
          & 94.30 & 5.28 & 83.74 & 97.04
          & 2008.7 & 293.4 & 1633.0 & 2400.2
          & 2681.6 & 347.8 & 2300.6 & 3169.6
          & 1.34 & 0.05 & 1.25 & 1.41 \\
        & FALCON\textsuperscript{\P}
          & 21.68 & 6.98 & 13.23 & 32.58
          & 747.0 & 73.4 & 629.2 & 899.0
          & 724.3 & 143.4 & 343.3 & 887.4
          & 0.96 & 0.14 & 0.53 & 1.14 \\
        & EPIC
          & \textbf{98.61} & 0.28 & 98.02 & 98.93
          & 726.3 & 41.1 & 618.9 & 793.6
          & 1864.8 & 96.7 & 1664.4 & 1977.9
          & 2.58 & 0.09 & 2.36 & 2.72 \\
        & \textbf{Ours}
          & 98.31 & 0.34 & 97.34 & 98.50
          & \textbf{664.7} & 39.3 & 607.4 & 723.9
          & \textbf{1759.0} & 86.2 & 1672.3 & 1927.9
          & \textbf{2.64} & 0.06 & 2.56 & 2.77 \\
        \midrule
        \multirow{4}{*}{Tunnel} & FUEL\textsuperscript{\P}
          & 59.71 & 16.09 & 37.93 & 76.31
          & 2400.2 & 0.0 & 2400.1 & 2400.2
          & 2748.1 & 937.3 & 1426.0 & 3491.6
          & 1.14 & 0.39 & 0.59 & 1.45 \\
        & FALCON\textsuperscript{\P}
          & 16.59 & 3.62 & 13.04 & 25.42
          & 462.8 & 257.8 & 115.0 & 879.8
          & 579.0 & 357.0 & 169.0 & 1174.4
          & 1.54 & 0.57 & 0.91 & 2.56 \\
        & EPIC\textsuperscript{\P}
          & 42.12 & 1.59 & 40.27 & 44.87
          & 718.5 & 36.9 & 661.3 & 777.6
          & 2328.9 & 111.6 & 2156.0 & 2495.9
          & 3.24 & 0.05 & 3.17 & 3.32 \\
        & \textbf{Ours}
          & \textbf{98.35} & 1.16 & 96.94 & 99.77
          & \textbf{1862.8} & 85.4 & 1680.2 & 1934.3
          & \textbf{5778.9} & 293.1 & 5162.0 & 6080.5
          & \textbf{3.10} & 0.06 & 2.99 & 3.16 \\
        \bottomrule
    \end{tabular}%
    }
\par\vspace{2pt}
\begin{minipage}{\textwidth}\scriptsize \textsuperscript{\P}Mean coverage is below 95\%; the reported time and path length therefore do not represent completion statistics at 95\% coverage. Bold values indicate the best mean performance among methods with mean coverage of at least 95\%.
\end{minipage}
\end{table*}

%% file: sections/conclusion.tex
\section{Conclusion}
\label{sec:conclusion}

We presented \method{}, a world-model-guided framework for LiDAR-based
UAV exploration. The world model learns how spatial observations evolve
with robot motion, using recurrent context to predict what candidate
actions may reveal beyond the current view. Predictive frontiers translate
these anticipated observations into spatial guidance for geometric
planning, allowing future sensing opportunities to inform exploration
decisions. New measurements update the model's predictions as the mission
progresses. Simulations and real-world experiments support improved
exploration performance and the feasibility of onboard inference.
This work establishes a way to integrate world model prediction into
autonomous exploration through predictive frontiers.

%% file: references.bib
@article{hafner2025dreamerv3,
  author  = {Danijar Hafner and Jurgis Pasukonis and Jimmy Ba and Timothy Lillicrap},
  title   = {Mastering Diverse Control Tasks through World Models},
  journal = {Nature},
  year    = {2025},
  volume  = {640},
  number  = {8059},
  pages   = {647--653},
  doi     = {10.1038/s41586-025-08744-2},
}

@inproceedings{romero2026dreamtofly,
  author    = {Angel Romero and Ashwin Shenai and Ismail Geles and Elie Aljalbout and Davide Scaramuzza},
  title     = {Dream to Fly: Model-Based Reinforcement Learning for Vision-Based Drone Flight},
  booktitle = {2026 IEEE International Conference on Robotics and Automation (ICRA)},
  year      = {2026},
  note      = {Accepted for publication},
  eprint    = {2501.14377},
  archiveprefix = {arXiv},
  primaryclass  = {cs.RO},
  url       = {https://arxiv.org/abs/2501.14377}
}

@misc{liu2026aird,
  author        = {Zian Liu and Andong Yang and Chunkai Yang and Ruidong An and Chao Gao and Guyue Zhou},
  title         = {{AirDreamer}: Generalist Drone Navigation with World Models},
  year          = {2026},
  eprint        = {2606.03252},
  archiveprefix = {arXiv},
  primaryclass  = {cs.RO},
  doi           = {10.48550/arXiv.2606.03252},
  url           = {https://arxiv.org/abs/2606.03252}
}

@misc{zhang2026mad,
  author        = {Xinhong Zhang and Runqing Wang and Yunfan Ren and Ding Yu and Boyu Zhou and Jian Sun and Fang Deng and Jie Chen and Gang Wang},
  title         = {{MAD}: Mapping-Aware World Models for Agile Quadrotor Flight},
  year          = {2026},
  eprint        = {2606.04534},
  archiveprefix = {arXiv},
  primaryclass  = {cs.RO},
  doi           = {10.48550/arXiv.2606.04534},
  url           = {https://arxiv.org/abs/2606.04534}
}

@article{zhou2021fuel,
  author  = {Boyu Zhou and Yichen Zhang and Xinyi Chen and Shaojie Shen},
  title   = {{FUEL}: Fast {UAV} Exploration Using Incremental Frontier Structure and Hierarchical Planning},
  journal = {IEEE Robotics and Automation Letters},
  year    = {2021},
  volume  = {6},
  number  = {2},
  pages   = {779--786},
  doi     = {10.1109/LRA.2021.3051563},
}

@article{geng2025epic,
  author  = {Shuang Geng and Zelin Ning and Fu Zhang and Boyu Zhou},
  title   = {{EPIC}: A Lightweight {LiDAR}-Based {AAV} Exploration Framework for Large-Scale Scenarios},
  journal = {IEEE Robotics and Automation Letters},
  year    = {2025},
  volume  = {10},
  number  = {5},
  pages   = {5090--5097},
  doi     = {10.1109/LRA.2025.3555878},
}

@inproceedings{tao2023seer,
  author    = {Yuezhan Tao and Yuwei Wu and Beiming Li and Fernando Cladera and Alex Zhou and Dinesh Thakur and Vijay Kumar},
  title     = {{SEER}: Safe Efficient Exploration for Aerial Robots Using Learning to Predict Information Gain},
  booktitle = {2023 IEEE International Conference on Robotics and Automation (ICRA)},
  year      = {2023},
  pages     = {1235--1241},
  doi       = {10.1109/ICRA48891.2023.10160295},
}

@inproceedings{ho2025mapex,
  author    = {Cherie Ho and Seungchan Kim and Brady Moon and Aditya Parandekar and Narek Harutyunyan and Chen Wang and Katia Sycara and Graeme Best and Sebastian Scherer},
  title     = {{MapEx}: Indoor Structure Exploration with Probabilistic Information Gain from Global Map Predictions},
  booktitle = {2025 IEEE International Conference on Robotics and Automation (ICRA)},
  year      = {2025},
  pages     = {13074--13080},
  doi       = {10.1109/ICRA55743.2025.11128862},
}

@misc{wang2026flowpilot,
  author        = {Runqing Wang and Ding Yu and Pengyuan Min and Xinhong Zhang and Wei Xiao and Yu Hu and Jie Chen and Fu Zhang and Gang Wang},
  title         = {{FlowPilot}: Real-Time World-Action Modeling for Agile {UAV} Navigation},
  year          = {2026},
  eprint        = {2608.00635},
  archiveprefix = {arXiv},
  primaryclass  = {cs.RO},
  doi           = {10.48550/arXiv.2608.00635},
  url           = {https://arxiv.org/abs/2608.00635}
}

@inproceedings{baek2025pipe,
  author    = {Seungjae Baek and Brady Moon and Seungchan Kim and Muqing Cao and Cherie Ho and Sebastian Scherer and Jeong Hwan Jeon},
  title     = {{PIPE} Planner: Pathwise Information Gain with Map Predictions for Indoor Robot Exploration},
  booktitle = {2025 IEEE/RSJ International Conference on Intelligent Robots and Systems (IROS)},
  year      = {2025},
  pages     = {7684--7691},
  doi       = {10.1109/IROS60139.2025.11246190},
  url       = {https://arxiv.org/abs/2503.07504}
}

@article{kong2023marsim,
  author  = {Fanze Kong and Xiyuan Liu and Benxu Tang and Jiarong Lin and Yunfan Ren and Yixi Cai and Fangcheng Zhu and Nan Chen and Fu Zhang},
  title   = {{MARSIM}: A Light-Weight Point-Realistic Simulator for {LiDAR}-Based {UAVs}},
  journal = {IEEE Robotics and Automation Letters},
  volume  = {8},
  number  = {5},
  pages   = {2954--2961},
  year    = {2023},
  doi     = {10.1109/LRA.2023.3264163}
}

@article{lambert2019lowlevel,
  author  = {Lambert, Nathan O. and Drew, Daniel S. and Yaconelli, Joseph and
             Levine, Sergey and Calandra, Roberto and Pister, Kristofer S. J.},
  title   = {Low-Level Control of a Quadrotor with Deep Model-Based
             Reinforcement Learning},
  journal = {IEEE Robotics and Automation Letters},
  year    = {2019},
  volume  = {4},
  number  = {4},
  pages   = {4224--4230},
  doi     = {10.1109/LRA.2019.2930489}
}

@misc{becker2020learning,
  author        = {Becker-Ehmck, Philip and Karl, Maximilian and Peters, Jan and
                   van der Smagt, Patrick},
  title         = {Learning to Fly via Deep Model-Based Reinforcement Learning},
  year          = {2020},
  eprint        = {2003.08876},
  archivePrefix = {arXiv},
  primaryClass  = {cs.RO},
  doi           = {10.48550/arXiv.2003.08876},
  url           = {https://arxiv.org/abs/2003.08876}
}

@misc{verraest2025skydreamer,
  author        = {Verraest, Aderik and Bahnam, Stavrow and Ferede, Robin and
                   de Croon, Guido and De Wagter, Christophe},
  title         = {{SkyDreamer}: Interpretable End-to-End Vision-Based Drone
                   Racing with Model-Based Reinforcement Learning},
  year          = {2025},
  eprint        = {2510.14783},
  archivePrefix = {arXiv},
  primaryClass  = {cs.RO},
  doi           = {10.48550/arXiv.2510.14783},
  url           = {https://arxiv.org/abs/2510.14783}
}

@inproceedings{bircher2016nbvp,
  author    = {Bircher, Andreas and Kamel, Mina and Alexis, Kostas and
               Oleynikova, Helen and Siegwart, Roland},
  title     = {Receding Horizon ``Next-Best-View'' Planner for {3D} Exploration},
  booktitle = {2016 IEEE International Conference on Robotics and Automation
               (ICRA)},
  year      = {2016},
  pages     = {1462--1468},
  doi       = {10.1109/ICRA.2016.7487281}
}

@inproceedings{cieslewski2017rapid,
  author    = {Cieslewski, Titus and Kaufmann, Elia and Scaramuzza, Davide},
  title     = {Rapid Exploration with Multi-Rotors: A Frontier Selection Method
               for High Speed Flight},
  booktitle = {2017 IEEE/RSJ International Conference on Intelligent Robots and
               Systems (IROS)},
  year      = {2017},
  pages     = {2135--2142},
  doi       = {10.1109/IROS.2017.8206030}
}

@inproceedings{witting2018history,
  author    = {Witting, Christian and Fehr, Marius and B{\"a}hnemann, Rik and
               Oleynikova, Helen and Siegwart, Roland},
  title     = {History-Aware Autonomous Exploration in Confined Environments
               Using {MAV}s},
  booktitle = {2018 IEEE/RSJ International Conference on Intelligent Robots and
               Systems (IROS)},
  year      = {2018},
  pages     = {1--9},
  doi       = {10.1109/IROS.2018.8594502}
}

@article{selin2019aep,
  author  = {Selin, Magnus and Tiger, Mattias and Duberg, Daniel and
             Heintz, Fredrik and Jensfelt, Patric},
  title   = {Efficient Autonomous Exploration Planning of Large-Scale
             {3-D} Environments},
  journal = {IEEE Robotics and Automation Letters},
  year    = {2019},
  volume  = {4},
  number  = {2},
  pages   = {1699--1706},
  doi     = {10.1109/LRA.2019.2897343}
}

@inproceedings{dai2020fastfrontier,
  author    = {Dai, Anna and Papatheodorou, Sotiris and Funk, Nils and
               Tzoumanikas, Dimos and Leutenegger, Stefan},
  title     = {Fast Frontier-Based Information-Driven Autonomous Exploration
               with an {MAV}},
  booktitle = {2020 IEEE International Conference on Robotics and Automation
               (ICRA)},
  year      = {2020},
  pages     = {9570--9576},
  doi       = {10.1109/ICRA40945.2020.9196707}
}

@article{schmid2020informative,
  author  = {Schmid, Lukas and Pantic, Michael and Khanna, Raghav and
             Ott, Lionel and Siegwart, Roland and Nieto, Juan},
  title   = {An Efficient Sampling-Based Method for Online Informative Path
             Planning in Unknown Environments},
  journal = {IEEE Robotics and Automation Letters},
  year    = {2020},
  volume  = {5},
  number  = {2},
  pages   = {1500--1507},
  doi     = {10.1109/LRA.2020.2969191}
}

@article{dang2020graph,
  author  = {Dang, Tung and Tranzatto, Marco and Khattak, Shehryar and
             Mascarich, Frank and Alexis, Kostas and Hutter, Marco},
  title   = {Graph-Based Subterranean Exploration Path Planning Using Aerial
             and Legged Robots},
  journal = {Journal of Field Robotics},
  year    = {2020},
  volume  = {37},
  number  = {8},
  pages   = {1363--1388},
  doi     = {10.1002/rob.21993}
}

@inproceedings{tang2023bubble,
  author    = {Tang, Benxu and Ren, Yunfan and Zhu, Fangcheng and He, Rui and
               Liang, Siqi and Kong, Fanze and Zhang, Fu},
  title     = {Bubble Explorer: Fast {UAV} Exploration in Large-Scale and
               Cluttered {3D}-Environments Using Occlusion-Free Spheres},
  booktitle = {2023 IEEE/RSJ International Conference on Intelligent Robots and
               Systems (IROS)},
  year      = {2023},
  pages     = {1118--1125},
  doi       = {10.1109/IROS55552.2023.10342348}
}

@article{lindqvist2024nbt,
  author  = {Lindqvist, Bj{\"o}rn and Patel, Akash and L{\"o}fgren, Kalle and
             Nikolakopoulos, George},
  title   = {A Tree-Based Next-Best-Trajectory Method for {3-D} {UAV} Exploration},
  journal = {IEEE Transactions on Robotics},
  year    = {2024},
  volume  = {40},
  pages   = {3496--3513},
  doi     = {10.1109/TRO.2024.3422052}
}

@article{zhang2025falcon,
  author  = {Zhang, Yichen and Chen, Xinyi and Feng, Chen and Zhou, Boyu and
             Shen, Shaojie},
  title   = {{FALCON}: Fast Autonomous Aerial Exploration Using Coverage Path
             Guidance},
  journal = {IEEE Transactions on Robotics},
  year    = {2025},
  volume  = {41},
  pages   = {1365--1385},
  doi     = {10.1109/TRO.2024.3522148}
}

@inproceedings{shrestha2019map,
  author    = {Shrestha, Rakesh and Tian, Fei-Peng and Feng, Wei and
               Tan, Ping and Vaughan, Richard},
  title     = {Learned Map Prediction for Enhanced Mobile Robot Exploration},
  booktitle = {2019 International Conference on Robotics and Automation
               (ICRA)},
  year      = {2019},
  pages     = {1197--1204},
  doi       = {10.1109/ICRA.2019.8793769}
}

@inproceedings{feng2023predrecon,
  author    = {Feng, Chen and Li, Haojia and Gao, Fei and Zhou, Boyu and
               Shen, Shaojie},
  title     = {{PredRecon}: A Prediction-Boosted Planning Framework for Fast
               and High-Quality Autonomous Aerial Reconstruction},
  booktitle = {2023 IEEE International Conference on Robotics and Automation
               (ICRA)},
  year      = {2023},
  pages     = {1207--1213},
  doi       = {10.1109/ICRA48891.2023.10160933}
}
